\documentclass{bmvc2k}

\title{TAPM-Net: Trajectory-Aware Perturbation Modeling for Infrared Small Target Detection}

\addauthor{Hongyang Xie}{Hongyang.Xie@warwick.ac.uk}{1}
\addauthor{Hongyang He}{Hongyang.He@warwick.ac.uk}{1}
\addauthor{Victor Sanchez}{v.f.sanchez-silva@warwick.ac.uk}{1}

\addinstitution{
 Signal and Information Processing (SIP) Lab\\
 The University of Warwick\\
 Coventry, UK
}

\runninghead{Xie et al.}{TAPM-Net}

\usepackage{array}
\usepackage{amsfonts}
\usepackage{booktabs}
\usepackage[table]{xcolor}
\usepackage{graphicx}  
\usepackage{wrapfig}
\usepackage{soul}
\usepackage{float}
\usepackage{dblfloatfix}

\renewcommand{\arraystretch}{1.2}

\begin{document}

\maketitle

\begin{abstract}
Infrared small target detection (ISTD) remains a long-standing challenge due to weak signal contrast, limited spatial extent, and cluttered backgrounds. Despite performance improvements from convolutional neural networks (CNNs) and Vision Transformers (ViTs), current models lack the mechanism to trace how small targets trigger directional, layer-wise perturbations in the feature space—an essential cue for distinguishing signal from structured noise in infrared scenes. To address this limitation, we propose the Trajectory-Aware Mamba Propagation Network (TAPM-Net) that explicitly models the spatial diffusion behavior of target-induced feature disturbances. TAPM-Net is built upon two novel components: a Perturbation-guided Path Module (PGM) module  and a Trajectory-Aware State Block (TASB). The PGM module constructs perturbation energy fields from multi-level features and extracts gradient-following feature trajectories that reflect the directionality of local responses. The resulting feature trajectories are fed into the TASB, which is a Mamba-based state-space unit that models dynamic propagation along each trajectory while incorporating velocity-constrained diffusion and semantic-aligned feature fusion from word- and sentence-level embeddings. Unlike existing attention-based methods, TAPM-Net enables anisotropic, context-sensitive state transitions along spatial trajectories, maintaining global coherence at low computational cost. Experiments on NUAA-SIRST and IRSTD-1K demonstrate that TAPM-Net achieves SOTA performance in ISTD.
\end{abstract}

%-------------------------------------------------------------------------
\section{Introduction}
Infrared small target detection (ISTD) is a crucial task in surveillance, remote sensing, and defense applications. %, where identifying minute targets in cluttered, low-contrast scenes remains a formidable challenge. T
Traditional approaches for ISTD predominantly focus on background suppression~\cite{Author20,Author21,Author22,Author23}, and low-rank~\cite{author24,Author25} and sparse decomposition techniques~\cite{Author26,Author27,Author3}. These approaches aim to isolate targets by modeling the background as a low-rank structure while treating the targets as sparse outliers. However, they often suffer from low detection accuracy, poor generalization to complex scenes, and vulnerability to structured noise.

% 放在一个较长段落的开头，前后都不要空行
%\noindent

\begin{wrapfigure}[21]{l}{0.55\columnwidth}% 
%\begin{figure}[t]{0.55\columnwidth}% 
  \vspace{8pt}
  \raggedright
  \includegraphics[width=0.55\textwidth]{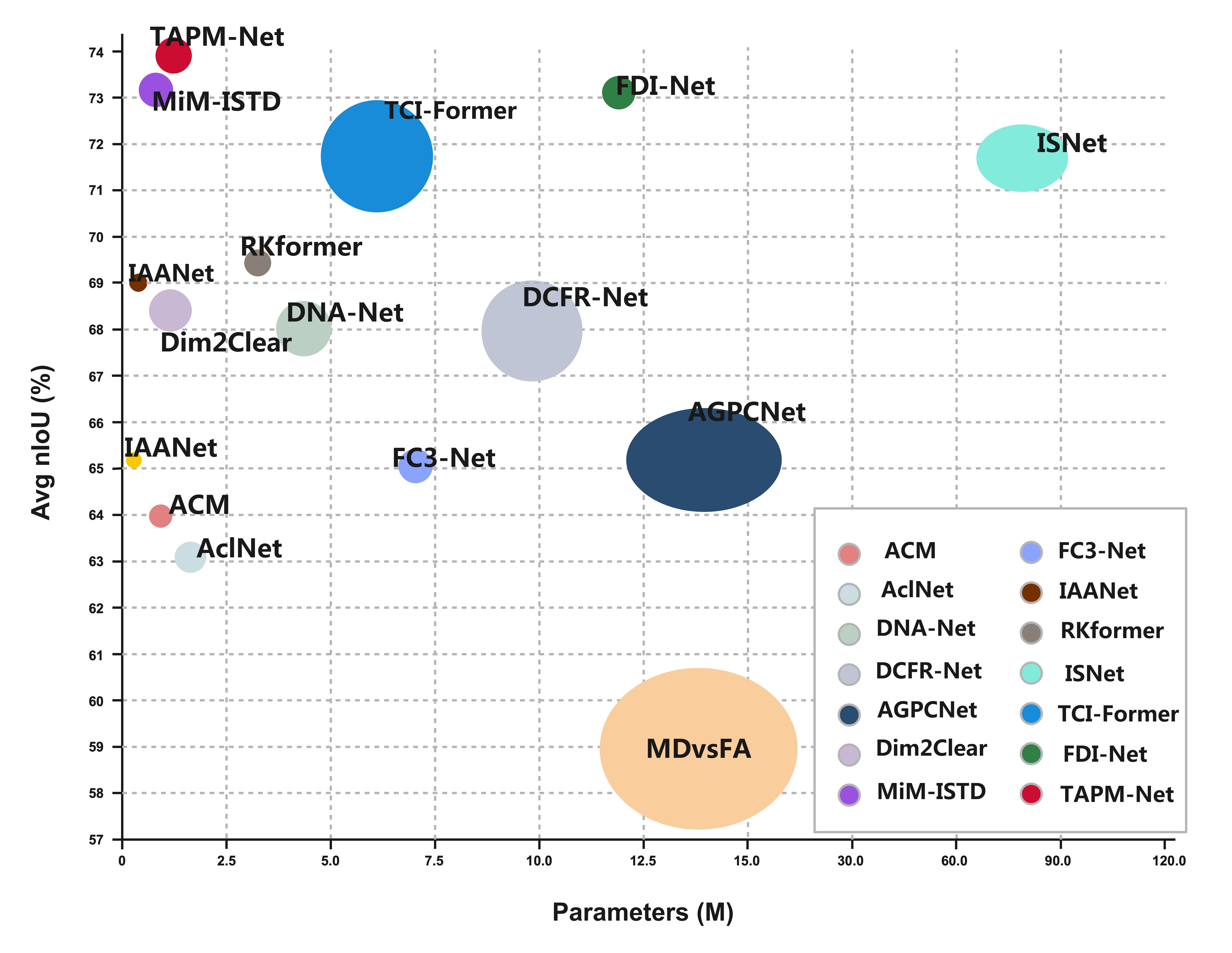}
  %scale=0.17
  \vspace{-25pt}
  \caption{Performance vs. complexity of several models on the NUAA-SIRST dataset. %The horizontal axis represents the number of model parameters (in millions), and the vertical axis shows the average nIoU(\%). 
  %Different colors represent different 
  Each model is represented by a distinct color. %detection methods, and 
  The size of each marker reflects the relative computational cost of the corresponding model.}
  \label{fig1}
  \vspace{-5cm}
\end{wrapfigure}
\vspace{-50pt}
\noindent

By learning data-driven discriminative features, CNN-based methods~\cite{Author5,Author6,Author7,Author8,Author9,Author15,Author29,Author30,Author31,Author32,Author33,Author40,Author34,Author35} outperform traditional ISTD methods in both precision and recall. However, many of these methods still rely on local contrast cues or global saliency estimation, and thus struggle to maintain fine-grained sensitivity to subtle, dispersed targets, particularly in highly textured or dynamic environments. Recently, ViTs~\cite{Author16} and hybrid CNN-Transformer architectures~\cite{Author13,Author14,Author17} have %emerged as powerful alternatives, 
further pushed detection accuracy by leveraging global self-attention and cross-layer fusion. Despite their strong performance, these models are typically computationally expensive, making them impractical for resource-constrained applications. Moreover, they lack an explicit mechanism to trace how feature disturbances propagate spatially, especially when such disturbances are weak or distributed.

To address these limitations, we draw inspiration from physical diffusion models~\cite{Author19,Author36,Author37,Author38,Author39} and propose a fundamentally different approach: treating small targets not as isolated saliency peaks but as sources of structured perturbation that propagate through the feature space. We then introduce TAPM-Net, a Trajectory-Aware Mamba Propagation Network that models the spatial dynamics of target-induced disturbances in a principled and learnable manner. TAPM-Net comprises two key modules: the Perturbation-guided Path Module (PGM), which constructs energy fields from backbone features and extracts gradient-aligned propagation feature trajectories; and the Trajectory-Aware State Block (TASB), which uses the Mamba architecture to recursively model state transitions along each trajectory. This design enables efficient, directional, and semantically aligned propagation modeling with low computational overhead. Unlike attention-based models that operate globally and uniformly, TAPM-Net focuses on structured, trajectory-aware propagation, thus capturing not only where the target is, but how its influence spreads. As shown in Figure~\ref{fig1}, TAPM-Net outperforms existing methods in both accuracy and efficiency by fully leveraging data-driven spatial propagation. Instead of relying on handcrafted priors, TAPM-Net learns perturbation-aware feature trajectories directly from feature distributions and models them efficiently using a Mamba-based state-space formulation. Our contributions are then threefold:
\begin{itemize}
    \item Our work is the first to introduce a perturbation-based formulation into the ISTD task, modeling small targets as localized disturbance sources and constructing energy fields to trace their spatial influence.
    
    \item Inspired by physical diffusion, we design a trajectory-aware state-space modeling framework built on Mamba, enabling efficient and semantically aligned feature propagation along gradient-guided trajectories.
    
    \item TAPM-Net achieves superior accuracy and computational efficiency in the NUAA-SIRST and IRSTD-1K datasets, confirming its effectiveness under both detection and deployment constraints.
\end{itemize}

\section{Related work}
\textbf{ISTD networks.}
Recent advances in ISTD have been driven by deep learning models, particularly CNNs, evaluated on the NUAA-SIRST~\cite{Author6} benchmark, first introduced by ACM~\cite{Author6}  to promote standardized evaluation. For example, MDvsFA~\cite{Author5} proposes feature aggregation to balance false alarms and miss-detections. DNANet~\cite{Author8} introduces dense nested connections for multi-scale context modeling, while UIUNet~\cite{Author41} leverages a U-Net to enhance spatial detail preservation. However, CNN-based methods typically focus on local feature extraction, often struggling to model weak target responses under strong background interference or to capture global context dependencies. To address these limitations, transformer-based and hybrid models have emerged. For example, TCI-Former~\cite{Author17} introduces target-contour interaction to jointly model target structures and their surrounding context, improving boundary preservation. MiM-ISTD~\cite{Author18}  proposes a nested Mamba-in-Mamba architecture to model feature perturbation propagation across scales, enhancing robustness in complex scenes. These methods partially bridge local-global feature integration but remain computationally demanding.

Despite recent advances, ISTD still faces long-term challenges in robustly modeling weak, structureless targets in cluttered environments while maintaining computational efficiency. Motivated by these gaps, our proposed TAPM-Net introduces trajectory-guided feature modeling and Mamba-based state propagation, enabling efficient disturbance-aware detection with enhanced spatial coherence and long-range dependency modeling.

\textbf{State-space sequence modeling in vision tasks.}
These models emerge as efficient alternatives to self-attention mechanisms to capture long-range dependencies in vision tasks. For example, U-Mamba~\cite{Author42}  extends Mamba to dense pixel-wise prediction by embedding state transitions into a U-Net, improving spatial detail retention. Swin-UNet~\cite{Author43} integrates shifted window mechanisms with hierarchical U-Net decoding, enhancing local-global feature alignment. Weak-Mamba-U-Net~\cite{Author44} introduces lightweight state updates to reduce computational overhead while preserving temporal dynamics. LocalVMamba~\cite{Author45} constrains state propagation within local regions, improving locality awareness but limiting global context modeling. MS-VMamba~\cite{Author46} extends Mamba to multi-scale feature processing, enabling cross-scale dependency learning. GroupMamba~\cite{Author47} further decomposes state transitions into grouped channels, improving efficiency in large-scale vision tasks. While these designs demonstrate the versatility of state-space modeling, they often rely on spatially uniform token propagation, lacking physically grounded mechanisms to guide feature flows based on disturbance structures. Mamba’s linear-time recurrence offers clear advantages over attention mechanisms in terms of scalability and efficiency, but current formulations treat all spatial positions equally, ignoring localized perturbation dynamics critical to ISTD. To address this, TAPM-Net integrates trajectory-guided state propagation into the Mamba framework, enabling disturbance-aware modeling that explicitly captures anisotropic feature flow patterns induced by small targets, without sacrificing computational efficiency.

\section{Proposed method}

\begin{figure}[t]
  \centering
  \includegraphics[width=\textwidth]{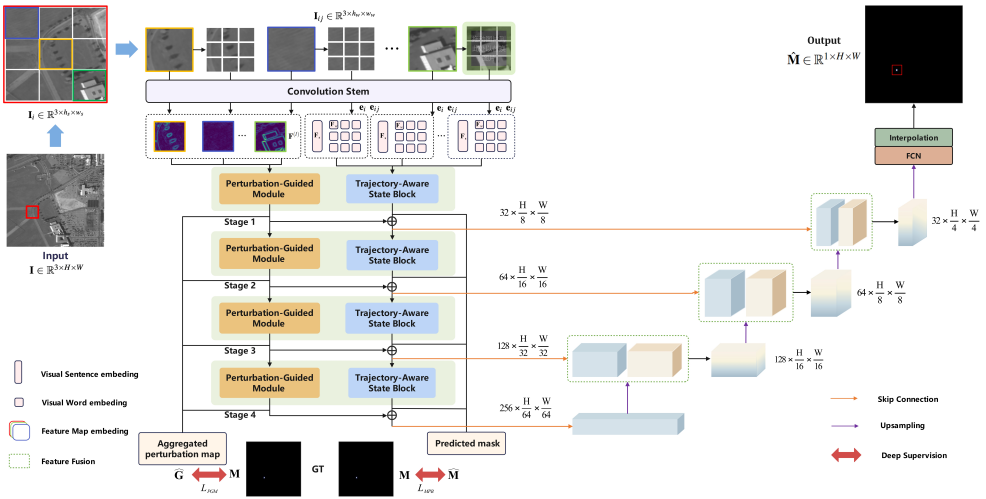}
  \vspace{-0.5cm}
  \caption{TAPM-Net architecture. It consists of a convolutional stem that extracts visual sentence and word embeddings, followed by multi-stage PGM and TASB for feature enhancement. Multi-scale features are fused through skip connections and upsampled during decoding to produce the final detection mask.}
  \label{fig2}
\end{figure}

%\subsection{Overview}
{TAPM-Net} is an ISTD solution that integrates a U-Net encoder-decoder with Mamba-based state-space modeling. As shown in Figure~\ref{fig2}, its architecture consists of two key components: hierarchical visual feature modeling, i.e., the PGM, and the TASB. TAPM-Net first segments the input image into multi-level visual regions to capture both local and contextual information. These features are encoded using a Mamba-based multi-scale encoder, followed by perturbation-aware trajectory extraction based on local energy fields. The extracted feature trajectories are then dynamically modeled and aligned using state-space mechanisms. Finally, the decoder computes the final prediction in the form of a mask.

Given an input image $\mathbf{I} \in \mathbb{R}^{3 \times H \times W}$, TAPM-Net first divides it into $n$ non-overlapping regions, each denoted by $\mathbf{I}_i \in \mathbb{R}^{3 \times h_s \times w_s}$. These regions are called visual sentences, with $h_s = H / \sqrt{n}$ and $w_s = W / \sqrt{n}$. Each sentence is further divided into $m$ sub-patches, termed visual words, each denoted by $\mathbf{I}_{ij} \in \mathbb{R}^{3 \times h_w \times w_w}$,  where $h_w = h_s / \sqrt{m}$ and $w_w = w_s / \sqrt{m}$. A lightweight convolutional stem (3$\times$3 Conv, BatchNorm, GELU) extracts word embeddings, each denoted by $\mathbf{e}_{ij} \in \mathbb{R}^{C_w}$, which are aggregated into word-level features, $\mathbf{F}_w \in \mathbb{R}^{n \times m \times C_w}$. Sentence-level embeddings, each denoted by $\mathbf{e}_i \in \mathbb{R}^{C_s}$, are computed by pooling word features, forming $\mathbf{F}_s \in \mathbb{R}^{n \times C_s}$. This hierarchical representation provides a semantic basis for the subsequent perturbation and trajectory modeling stages.

\begin{figure}[t]
  \centering
  \includegraphics[width=\textwidth]{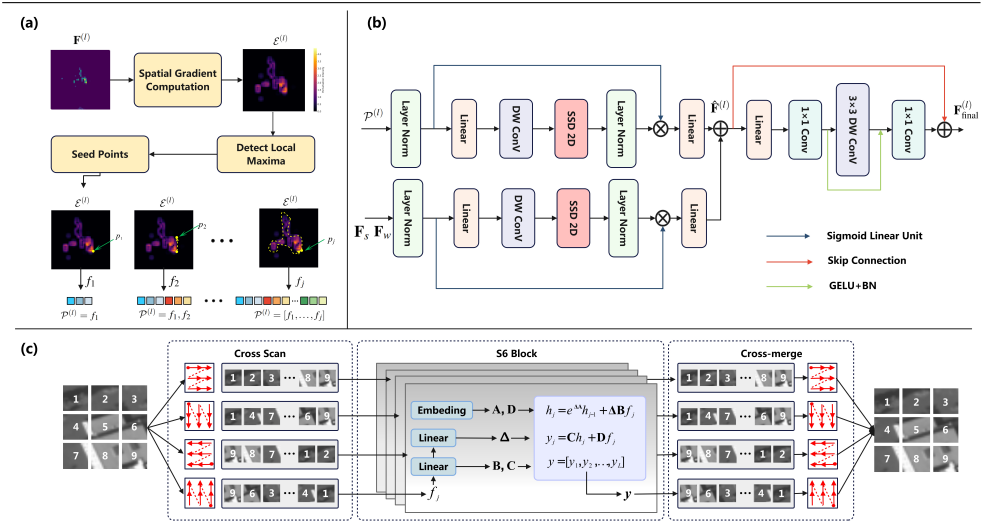} 
   \vspace{-0.5cm}
  \caption{(a) PGM extracts trajectories from spatial energy maps. (b) TASB models feature dynamics along trajectories. (c) The SS2D block enhances spatial dependency through cross-scan and cross-merge operations.}
  \label{fig3}
\end{figure}

\textbf {PGM.} As shown in Figure \ref{fig3} (a), at each stage $l \in {[1,2,3,4]}$, the backbone feature map, \( \mathbf{F}^{(l)} \in \mathbb{R}^{C_l \times H_l \times W_l} \), is passed to the PGM. The feature map  $\mathbf{F}^{(l)}$, constructed by a sequence of Mamba blocks, encodes rich semantic representations while retaining sufficient spatial resolution for precise localization of small objects. To explicitly characterize local disturbances potentially caused by small targets, we construct scalar-valued perturbation energy maps, each denoted by \( \mathcal{E}^{(l)} \in \mathbb{R}^{H_l \times W_l} \). These maps are computed by estimating spatial gradients across the feature channels and accumulating the magnitude of changes at each location. Specifically, the energy at each position $(x,y)$ %of the backbone feature map \( \mathbf{F}^{(l)}\) 
is computed as:
\begin{equation}
\mathcal{E}^{(l)}(x, y) = \sum_{c=1}^{C_l} \left| \mathbf{F}_c^{(l)}(x+1, y) - \mathbf{F}_c^{(l)}(x-1, y) \right| + \left| \mathbf{F}_c^{(l)}(x, y+1) - \mathbf{F}_c^{(l)}(x, y-1) \right|,
\end{equation}
\noindent where %\( (x, y) \) indexes a spatial location, and 
\( \mathbf{F}_c^{(l)} \) denotes the \( c \)-th channel slice of $\mathbf{F}^{(l)}$. This operation aggregates directional differences across channels, highlighting locations with strong spatial discontinuities such as edges or localized perturbations, often associated with the presence of the target.

We initiate trajectory sampling from local maxima in each energy map, $\mathcal{E}^{(l)}$. Starting from a seed point \( p_1 \in \mathbb{R}^2 \), we follow the direction of steepest gradient to generate a trajectory of sampled positions. The trajectory is recursively updated by:
\begin{equation}
p_{j+1} = p_j + \eta \cdot \frac{\nabla \mathcal{E}^{(l)}(p_j)}{\| \nabla \mathcal{E}^{(l)}(p_j) \|_2 + \epsilon},
\end{equation}
\noindent where \( \eta \) is a step-size factor controlling the spatial resolution of the trace, and \( \epsilon \) is a small constant for numerical stability. The gradient \( \nabla \mathcal{E}^{(l)} \) is computed via finite differences or Sobel filters. The length of each trajectory, denoted by \( L \), can be either fixed or adaptively determined based on the energy decay.
Each point $p_j$ along the trajectory is mapped back to the backbone feature space. Using bilinear interpolation, we extract channel-wise feature vectors; specifically,  the sampled positions $\{p_j\}$ along the traced perturbation trajectory on the energy map $\mathcal{E}^{(l)}$ % which correspond to spatial coordinates on the backbone feature map $\mathbf{F}^{(l)}$. 
are used to form a sequence of feature tokens:

\begin{equation}
f_j = \text{Interp}(\mathbf{F}^{(l)}, p_j), \quad \mathcal{P}^{(l)} = [f_1,f_2 \dots, f_L] \in \mathbb{R}^{L \times C_l},
\end{equation}

\noindent where \( f_j \in \mathbb{R}^{C_l} \) represents the local feature at point \( p_j \) and $\text{Interp}(\cdot,\cdot)$ denotes bilinear interpolation. The resulting sequence \( \mathcal{P}^{(l)} \) encodes the spatial dynamics of perturbation propagation and serves as input to the subsequent Mamba-based trajectory modeling module.

\textbf{TASB.} This block models the spatial propagation of perturbation patterns induced by small targets and aligns these dynamics with global context. As shown in Figure~\ref{fig3}(b) and (c), TASB is built on a Mamba-based Selective State-Space 2D Block (SS2D), here instantiated as the S6 Block. The S6 Block first applies cross-scan operations to unfold spatial features into directional sequences, models their sequential dependencies using a state-space update mechanism, and then recombines them via cross-merge operations to restore the spatial layout.

Starting with the sequence $\mathcal{P}^{(l)} = [f_1, \dots, f_L]$, TASB applies a state-space model to recursively update the hidden states. The update is formulated as:
\begin{equation}
h_j = e^{\Delta \mathbf{A}} h_{j-1} + \Delta \mathbf{B} f_j; \quad \textrm{ } \quad 
y_j = \mathbf{C} h_j + \mathbf{D} f_j; \quad
y = [y_1,y_2,...,y_j],
\end{equation}
%\begin{equation}
%\mathbf{y}_j = \mathbf{C} \mathbf{h}_j + \mathbf{D} f_j
%\end{equation}

\noindent where $h_j$ represents the hidden state at step $j$, $y_j$ denotes the corresponding output, $y$ is the collection of all state outputs %$y_j$ 
along the entire trajectory or sequence, $\mathbf{A}$ is the state transition matrix that controls how the hidden state evolves over the sequence, $\mathbf{B}$ projects feature $f_j$ into the state update, $\mathbf{C}$ maps the updated hidden state to the output space, $\mathbf{D}$ allows for a direct contribution of the input feature to the output, and $\Delta$ is a learnable step-size parameter that regulates the dynamics of the update.

SS2D includes a bottleneck structure, which first reduces the channel dimension using a $1 \times 1$ linear layer before the state update, and then restores the original dimension after the update. This design improves computational efficiency while preserving feature representation capacity. To enhance spatial modeling, TASB adopts a cross-scanning strategy, where the feature map $\hat{\mathbf{F}}^{(l)}$ [see Figure \ref{fig3}(b)]
is unfolded into four directional sequences: horizontal, vertical, diagonal, and anti-diagonal. These sequences are processed independently and merged back to restore the spatial layout.

After state modeling, TASB aligns the state output $y_j$ with the corresponding feature $f_j$, word-level feature $\mathbf{F}_w(x_j, y_j)$, and sentence-level feature $\mathbf{F}_s(x_j, y_j)$. The aligned feature is computed by concatenation followed by a linear projection, $\phi(\cdot)$:

\begin{equation}
\mathbf{z}_j = \phi \left( [\mathbf{y} || f_j || \mathbf{F}_w(x_j, y_j) || \mathbf{F}_s(x_j, y_j)] \right).
\end{equation}

TASB projects \( \mathbf{z}_j \); i.e., the aligned trajectory states, back to their original spatial positions \( (x_j, y_j) \) on the feature map $\hat{\mathbf{F}}^{(l)}$. When multiple states map to the same location, they are averaged by dividing by the number of contributions, \( N(x, y) \). The resulting aggregated feature map \( \hat{\mathbf{F}}^{(l)} \) is then fused with the original backbone feature \( \mathbf{F}^{(l)} \) through residual addition, scaled by a fusion weight \( \lambda \) to control the contribution of the trajectory-aware features. This multi-scale fusion process ensures that spatial details, semantic alignment, and perturbation dynamics are jointly captured and progressively enhanced across the feature hierarchy.

\textbf{Decoding and multi-scale feature fusion.} After the multi-stage feature enhancement, the fused feature map $\mathbf{F}_{\text{final}}^{(l)}$  is the trajectory-enhanced feature map obtained by fusing $\mathbf{F}^{(l)}$ with the TASB-enhanced map $\hat{\mathbf{F}}^{(l)}$ [see Figure \ref{fig3}(b)]:  
\begin{equation}
\mathbf{F}_{\text{final}}^{(l)} = \mathbf{F}^{(l)} + \lambda \cdot \hat{\mathbf{F}}^{(l)}.
\end{equation}
The decoder contains three upsampling stages, where each stage performs bilinear interpolation followed by a 3$\times$3 convolution, BatchNorm, and ReLU activation. The feature channels are gradually reduced from 256 to 32 (see Figure~\ref{fig2}). At each stage, skip connections concatenate the corresponding encoder features. Finally, a fully convolutional network (FCN) head projects the decoded features to a single-channel prediction, \( \hat{\mathbf{M}} \in \mathbb{R}^{1\times H \times W} \), followed by interpolation to match the original image size.

\textbf{Loss function.} TAPM-Net is optimized with a dual-branch loss. The main segmentation loss combines binary cross-entropy (BCE) and Dice loss:
%\begin{equation}
$\mathcal{L}_{\text{seg}} = \text{BCE}(\hat{\mathbf{M}}, \mathbf{M}) + \alpha \cdot \text{Dice}(\hat{\mathbf{M}}, \mathbf{M})$, 
%\end{equation}
where \( \hat{\mathbf{M}} \) is the predicted mask, \( \mathbf{M} \) is the ground truth, and \( \alpha \) balances the two terms. To guide learning, we aggregate all feature trajectorys back to their spatial locations, forming a perturbation response map $\hat{\mathbf{G}} \in \mathbb{R}^{H \times W}$. This map highlights regions influenced by perturbation flows and is used as auxiliary. : $\mathcal{L}_{\text{PGM}} = \text{BCE}(\hat{\mathbf{G}}, \mathbf{M})$. 
%\begin{equation}
%\mathcal{L}_{\text{PGM}} = \text{BCE}(\hat{\mathbf{G}}, \mathbf{M})
%\end{equation}
The total training objective is formulated as:
\begin{equation}
\mathcal{L}_{\text{total}} = \mathcal{L}_{\text{seg}} + \beta \cdot \mathcal{L}_{\text{PGM}},
\end{equation}
where \( \beta \) controls the contribution of the auxiliary perturbation supervision. This loss design ensures both accurate boundary prediction and spatial consistency.

\section{Performance evaluation}
%\subsection{Experimental Settings}
\textbf{Datasets.}
%To thoroughly evaluate the proposed method, 
We conduct experiments on two widely used ISTD datasets: NUAA-SIRST~\cite{Author6} and IRSTD-1k~\cite{Author15}. The NUAA-SIRST dataset contains 427 infrared images captured under diverse scenes, backgrounds, and target scales, providing rich variability for robust evaluation. The IRSTD-1k dataset includes 1000 images with a resolution of 512$\times$512 pixels, covering challenging scenarios such as cluttered backgrounds, low signal-to-noise ratios, and small targets. Following standard practice, we randomly split each dataset into 80\% for training and 20\% for testing.

\textbf{Evaluation Metrics.} We adopt both pixel-level and object-level metrics, consistent with standard ISTD benchmarks. Pixel-level metrics include Intersection over Union (IoU) and Normalized IoU (nIoU), which measure spatial overlap quality. IoU evaluates the ratio between the intersection and union of predicted and ground truth masks, while nIoU mitigates scale sensitivity across different target sizes. Object-level metrics include the Probability of Detection ($P_d$) and False Alarm Rate ($F_a$).

\textbf{Quantitative comparisons.} We compare TAPM-Net with traditional, CNN-based, ViT-based, and hybrid methods as tabulated in Table~\ref{tab1}. For pixel-level metrics (IoU and nIoU), hybrid methods outperform traditional methods, but many still struggle with predicting precise boundaries for small targets. TAPM-Net achieves the best scores by leveraging perturbation-guided and trajectory-aware modeling. For object-level metrics ($P_d$ and $F_a$), TAPM-Net balances detection and false alarms better than others, achieving perfect detection with low false alarms on NUAA-SIRST and maintaining strong performance on IRSTD-1k, thus showing superior robustness in challenging infrared scenarios.

\begin{table*}[!t]
\centering
\caption{Performance of several methods on NUAA-SIRST and IRSTD-1k. Results in \colorbox{red!30}{\hspace{0.5em}}, \colorbox{orange!25}{\hspace{0.5em}}, and \colorbox{orange!15}{\hspace{0.5em}} indicate, respectively, the best, second best, and third best performance.}
\renewcommand{\arraystretch}{1.2}
\resizebox{\textwidth}{!}{
\begin{tabular}{l@{\hspace{2mm}} | l | c  c c c c c c c}
\toprule
Method & Type & 
\multicolumn{4}{c}{NUAA-SIRST} & 
\multicolumn{4}{c}{IRSTD-1k} \\
& & IoU(\%)↑ & nIoU(\%)↑ & Pd(\%)↑ & Fa(\%)↓ & IoU(\%)↑ & nIoU(\%)↑ & Pd(\%)↑ & Fa(\%)↓ \\
\midrule
PSTNN~\cite{Author3} & Trad & 22.40 & 22.35 & 77.95 & 29.11 & 24.57 & 17.93 & 71.99 & 35.26 \\
MSLSTIPT~\cite{Author4} & Trad & 10.30 & 9.58 & 82.13 & 11.31 & 11.43 & 5.93 & 79.03 & 15.24 \\
NRAM~\cite{Author1} & Trad & 12.16 & 10.22 & 74.52 & 13.85 & 15.25 & 9.90 & 70.68 & 16.93 \\
TLLCM~\cite{Author2} & Trad & 1.03 & 0.91 & 79.09 & 58.99 & 3.31 & 0.78 & 77.39 & 67.38 \\
MDvsFA~\cite{Author5} & CNN & 60.30 & 58.26 & 89.35 & 56.35 & 49.50 & 47.41 & 82.11 & 80.33 \\
ACM~\cite{Author6} & CNN & 72.33 & 71.43 & 96.33 & 9.33 & 60.97 & 58.02 & 90.58 & 21.78 \\
AlcNet~\cite{Author7} & CNN & 74.31 & 73.12 & 97.34 & 20.21 & 62.05 & 59.58 & 92.19 & 31.56 \\
DNANet~\cite{Author8} & CNN & 75.27 & 73.68 & 98.17 & 13.62 & 69.01 & 66.22 & 91.92 & 17.57 \\
DCFR-Net~\cite{Author9} & CNN & 76.23 & 74.69 & 99.08 & 6.52 & 65.41 & 65.45 & 93.60 & \cellcolor{red!30} 7.35 \\
AGPCNet~\cite{Author10} & CNN & 70.60 & 70.16 & 97.25 & 37.44 & 62.82 & 63.01 & 90.57 & 29.82 \\
Dim2Clear~\cite{Author11} & CNN & 77.20 & 75.20 & 99.10 & 6.72 & 66.30 & 64.20 & 93.70 & 20.90 \\
FC3-Net~\cite{Author12} & CNN & 74.22 & 72.64 & 99.12 & 6.57 & 64.98 & 63.59 & 92.93 & 15.73 \\
IAANet~\cite{Author13} & CNN-ViT & 75.31 & 74.65 & 98.22 & 35.65 & 59.82 & 58.24 & 88.62 & 24.79 \\
RKformer~\cite{Author14} & CNN-ViT & 77.24 & 74.89 & 99.11 & \cellcolor{red!30} 1.58 & 64.12 & 64.18 & 93.27 & 18.65 \\
ISNet~\cite{Author15} & CNN & 80.02 & 78.12 & 98.84 & 4.92 & 68.77 & 64.84 & 95.56 & 15.39 \\
SegFormer~\cite{Author16} & ViT & 76.01 & 66.43 & 96.35 & 35.83 & 60.12 & 57.23 & 89.22 & 18.47 \\
TCI-Former~\cite{Author17} & CNN-ViT & 80.79 & 79.85 & \cellcolor{orange!15} 99.23 & 4.19 & 70.14 & 67.69 & 96.31 & 14.81 \\
MiM-ISTD~\cite{Author18} & Mamba & \cellcolor{orange!15} 80.92 & \cellcolor{orange!15} 80.13 & \cellcolor{orange!25} 100.00 & \cellcolor{orange!15} 2.17 & \cellcolor{orange!15} 70.36 & \cellcolor{orange!15} 68.05 & \cellcolor{orange!25} 96.95 & 13.38 \\
FDI-Net~\cite{Author19} & CNN-ViT & \cellcolor{orange!25} 81.86 & \cellcolor{orange!25} 81.08 & 99.14 & 4.51 & \cellcolor{orange!25} 72.01 & \cellcolor{orange!25} 71.99 & \cellcolor{red!30} 97.33 & \cellcolor{orange!15} 12.23 \\
\midrule
TAPM-Net (Ours) & Mamba & \cellcolor{red!30} 81.94 & \cellcolor{red!30} 81.24 & \cellcolor{red!30} 100.00 & \cellcolor{orange!25} 1.98 & \cellcolor{red!30} 74.38 & \cellcolor{red!30} 73.65 & \cellcolor{orange!15} 96.35 & \cellcolor{orange!25} 11.42 \\
\bottomrule
\end{tabular}
}
\label{tab1}
\end{table*}

 \begin{figure}[t]
  \centering
  \includegraphics[width=\textwidth]{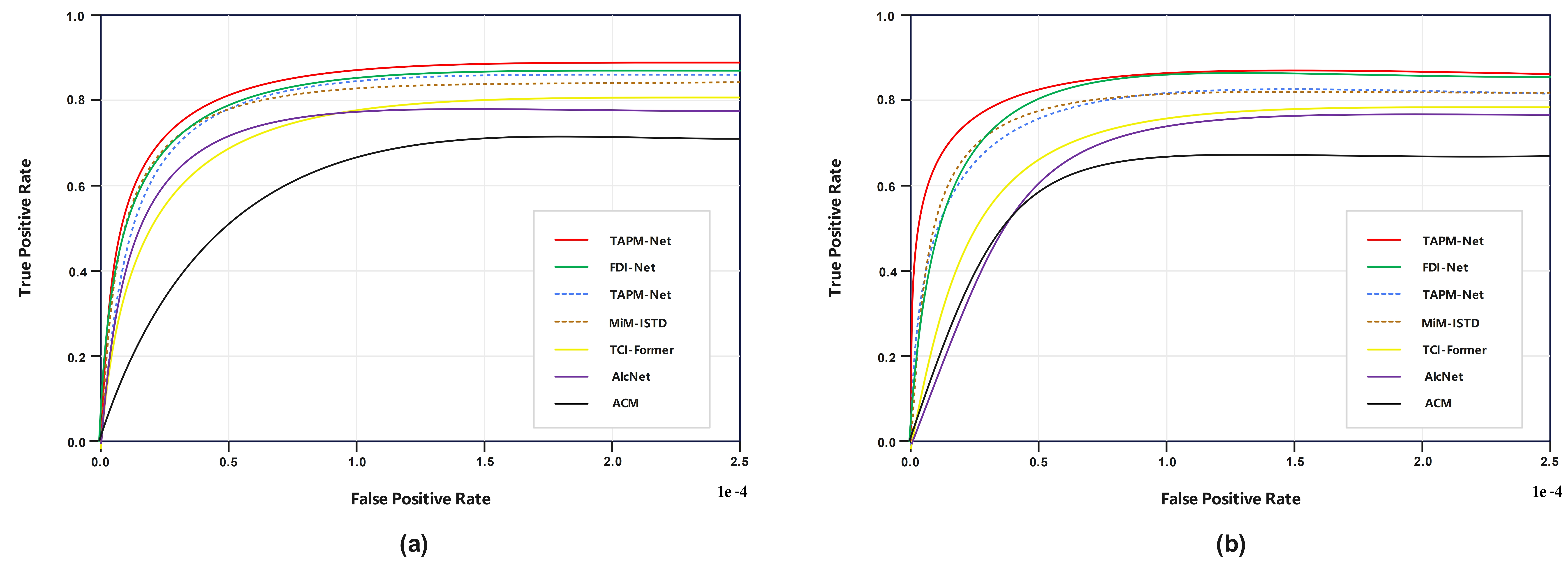} 
  \vspace{-1cm}
  \caption{ROC curves of several methods on (a)  NUAA-SIRST and (b) IRSTD-1k.}
  \label{fig4}
\end{figure}
%\subsection{Visual Results}

Figure~\ref{fig4} shows the true positive rate (TPR) and false positive rate (FPR) on NUAA-SIRST and IRSTD-1k. Hybrid models consistently achieve higher TPR at lower FPR compared to traditional approaches like ACM, demonstrating better sensitivity and false alarm control across both datasets. These results confirm the effectiveness of hybrid models for practical ISTD.

\begin{figure}[t]
  \centering
  \includegraphics[width=\textwidth]{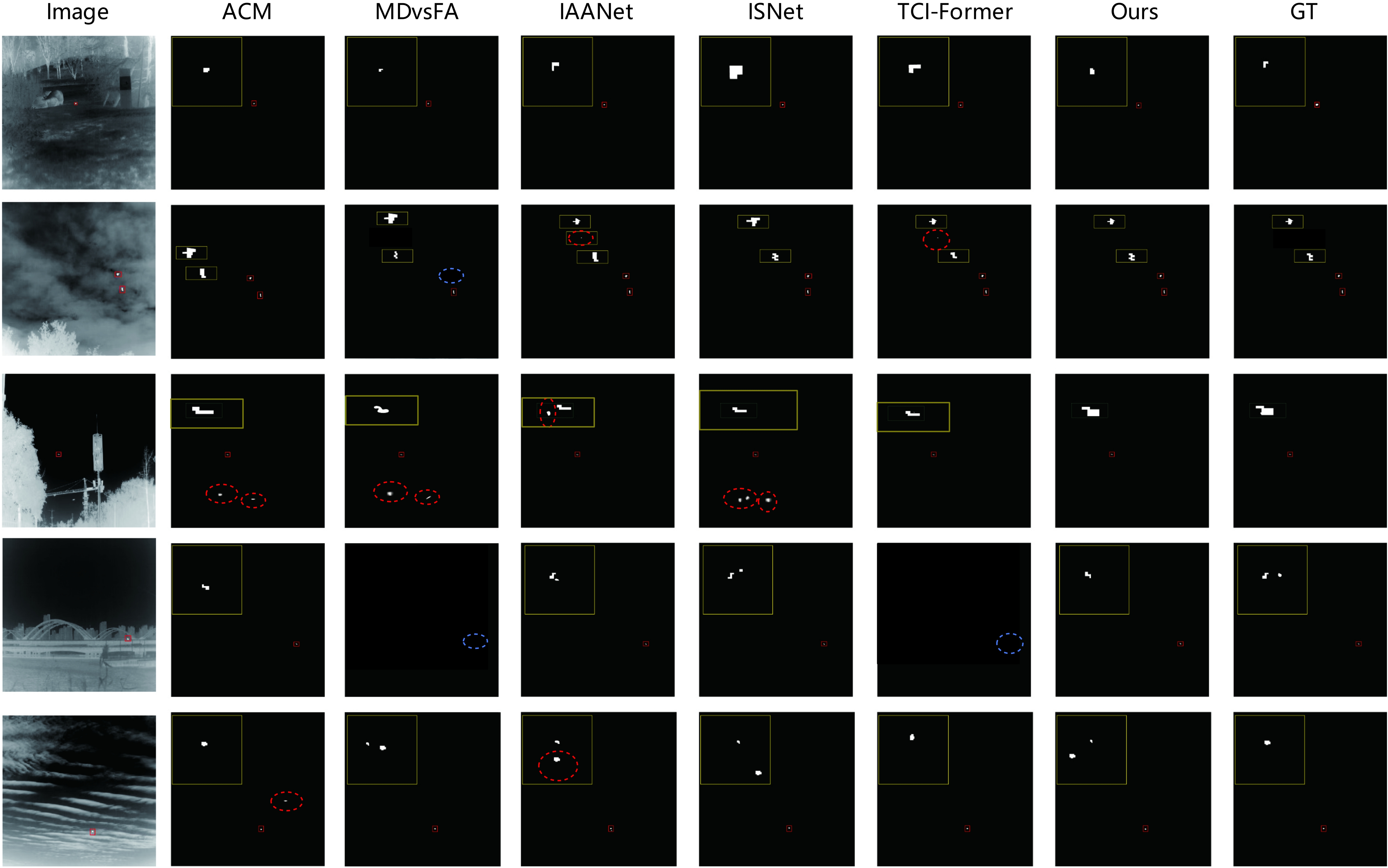} 
  \vspace{-0.7cm}
  \caption{Visualization of results of different ISTD methods, where the GT column depicts the ground truth. Red boxes indicate target locations, yellow boxes indicate magnified prediction regions, red circles mark false alarms, and blue circles indicate missed detections.}
  \label{fig5}
  \vspace{0.4cm}
\end{figure}

\textbf{Visualization of results.} Figure ~\ref{fig5} depicts visual results obtained by different methods. These results cover various scenarios such as background clutter, occlusion, and the presence of similar distracting objects. It can be observed that several methods tend to produce false alarms or miss targets in challenging cases. This highlights the inherent challenges of ISTD, particularly when targets are small, with low-contrast, or surrounded by confusing background structures.

\begin{figure}[t]
  \centering
  \includegraphics[width=\textwidth]{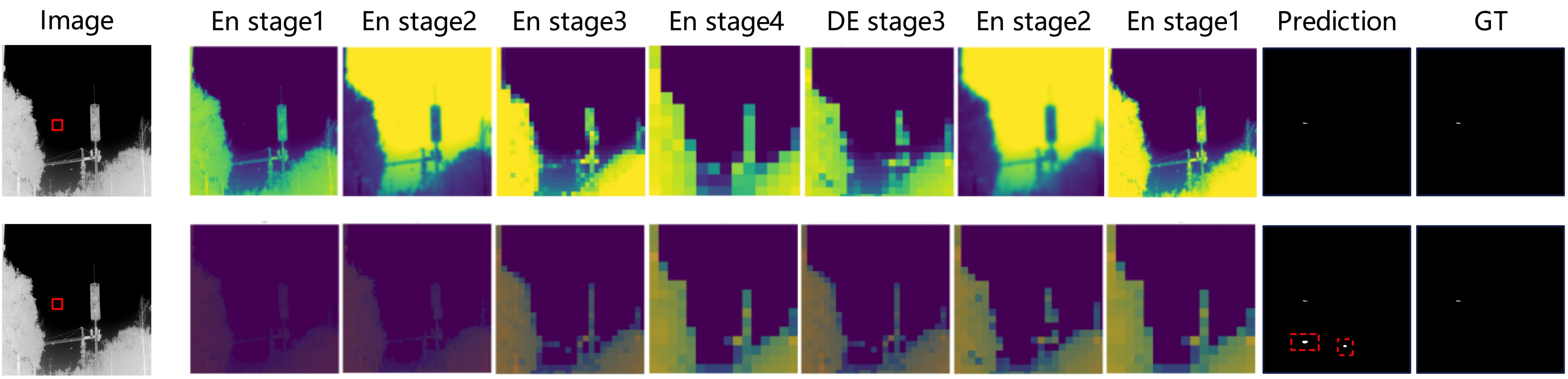} 
   \vspace{-0.7cm}
  \caption{Feature responses from the encoder (En-stage 1 to En-stage 4) and de-
coder (De-stage 3 to De-stage 1). The first row corresponds to TAPM-Net, while the second one to TCI-Former. The GT column depicts the ground truth.}
  \label{fig6}
\end{figure}

To analyze how TAPM-Net progressively refines target representations, Figure~\ref{fig6} depicts the multi-stage feature responses from the encoder (En-stage 1 to En-stage 4) and decoder (De-stage 3 to De-stage 1). The examples are selected from both NUAA-SIRST and IRSTD-1k datasets. As shown, the encoding stages gradually suppress background interference while enhancing the target response as the resolution decreases. The decoding stages progressively recover spatial details and refine the target regions towards the final prediction. These results show the complete transformation from raw inputs to features and then to accurate segmentation masks.

%\textbf{ROC curves}
%As shown in Figure~\ref{fig4}, we compare the true positive rate (TPR) and false positive rate (FPR) on NUAA-SIRST and IRSTD-1k. Learning-based methods consistently achieve higher TPR at lower FPR compared to traditional approaches like ACM, demonstrating better sensitivity and false alarm control across both datasets. These results confirm the effectiveness of learning-based models for practical infrared small target detection.

\textbf{Ablation study.} We assess the contribution of each component of TAPM-Net on NUAA-SIRST. % to using IoU, nIoU, Pd, and Fa. 
Table~\ref{tab2} shows that PGM and TASB independently improve detection, and their joint use achieves the best performance. The results in Table~\ref{tab3} further confirm the effectiveness of PGM’s explicit trajectory modeling and feature fusion. Finally, Table~\ref{tab4} shows that Mamba-based TASB outperforms standard residual and bottleneck blocks, demonstrating its advantage in capturing long-range spatial context and enhancing semantic consistency.
\begin{table*}[t]
\centering
\begin{minipage}[t]{0.48\textwidth}
\centering
\caption{Performance of TAPM-Net on NUAA-SIRST when different components are removed.}
\label{tab2}
\scriptsize
\begin{tabular}{@{}l|cccc@{}}
\toprule
Variant & IoU↑ & nIoU↑ & Pd↑ & Fa↓ \\
\midrule
U-Net & 63.12 & 66.56 & 91.23 & 43.23 \\
U-Net+PGM & 72.45 & 73.98 & 95.63 & 8.78 \\
U-Net+TASB & 74.98 & 76.12 & 98.26 & 6.45 \\
U-Net+PGM+TASB (full) & \textbf{81.04} & \textbf{80.24} & \textbf{100.0} & \textbf{1.98} \\
\bottomrule
\end{tabular}
\end{minipage}
\hfill
\begin{minipage}[t]{0.48\textwidth}
\centering
\caption{Performance of TAPM-Net on NUAA-SIRST with several variants of PGM.}
\label{tab3}
\scriptsize

\begin{tabular}{@{}l|cccc@{}}
\toprule
Variant & IoU↑ & nIoU↑ & Pd↑ & Fa↓ \\
\midrule
En. Map (only) & 50.44 & 52.92 & 88.23 & 19.23 \\
En. Map+Traj. & 74.78 & 76.52 & 98.45 & 7.18 \\
En. Map+Traj.+
\\Fusion (full) & \textbf{81.04} & \textbf{80.24} & \textbf{100.0} & \textbf{1.98} \\
\bottomrule
\end{tabular}

\end{minipage}

\vspace{1em} 

\begin{minipage}[t]{0.5\textwidth}
\centering
\caption{Performance of TAPM-Net on NUAA-SIRST with several variants of TASB.}
\label{tab4}
\scriptsize
\begin{tabular}{@{}l|cccc@{}}
\toprule
Variant & IoU↑ & nIoU↑ & Pd↑ & Fa↓ \\
\midrule
ResBlock & 79.12 & 79.88 & 98.49 & 4.51 \\
Bottle Neck & 78.89 & 79.36 & 98.95 & 5.14 \\
TASB (full) & \textbf{81.04} & \textbf{80.24} & \textbf{100.0} & \textbf{1.98} \\
\bottomrule
\end{tabular}
\end{minipage}
\end{table*}

\section{Conclusion}
In this work, we proposed TAPM-Net, a new ISTD method that combines perturbation-guided feature modeling with Mamba-based state-space learning. TAPM-Net introduces a two-step process that first builds energy maps to guide feature extraction along trajectories and then models feature propagation along these trajectories to enhance target representation. TAPM-Net further aligns these features with multi-level visual embeddings to improve spatial consistency. Experimental results on NUAA-SIRST and IRSTD-1k show that TAPM-Net achieves higher detection accuracy and lower false alarm rates compared to several methods, demonstrating its effectiveness for small target detection in complex infrared scenes.

%-------------------------------------------------------------------------

\bibliography{egbib}
\end{document}